\documentclass[sigconf]{acmart}

\copyrightyear{2022} 
\acmYear{2022} 
\setcopyright{acmlicensed}\acmConference[ICAIF '22]{3rd ACM International Conference on AI in Finance}{November 2--4, 2022}{New York, NY, USA}
\acmBooktitle{3rd ACM International Conference on AI in Finance (ICAIF '22), November 2--4, 2022, New York, NY, USA}
\acmPrice{15.00}
\acmDOI{10.1145/3533271.3561767}
\acmISBN{978-1-4503-9376-8/22/10}

\graphicspath{ {./figures/} }

\usepackage[sort]{cleveref}

\usepackage{csquotes}

\usepackage{mathtools}

\usepackage{booktabs}


\begin{document}

\title{Learning Not to Spoof}

\author{David Byrd}
\affiliation{%
  \institution{Bowdoin College}
  \country{United States}
}
\email{d.byrd@bowdoin.edu}

\begin{abstract}
As intelligent trading agents based on reinforcement learning (RL) gain prevalence, it becomes more important to ensure that RL agents obey laws, regulations, and human behavioral expectations.  There is substantial literature concerning the aversion of obvious catastrophes like crashing a helicopter or bankrupting a trading account, but little around the avoidance of subtle non-normative behavior for which there are examples, but no programmable definition.  Such behavior may violate legal or regulatory, rather than physical or monetary, constraints.

In this article, I consider a series of experiments in which an intelligent stock trading agent maximizes profit but may also inadvertently learn to spoof the market in which it participates.  I first inject a hand-coded spoofing agent to a multi-agent market simulation and learn to recognize spoofing activity sequences.  Then I replace the hand-coded spoofing trader with a simple profit-maximizing RL agent and observe that it independently discovers spoofing as the optimal strategy.  Finally, I introduce a method to incorporate the recognizer as normative guide, shaping the agent's perceived rewards and altering its selected actions.  The agent remains profitable while avoiding spoofing behaviors that would result in even higher profit.  After presenting the empirical results, I conclude with some recommendations.  The method should generalize to the reduction of any unwanted behavior for which a recognizer can be learned.
\end{abstract}


\maketitle

\section{Introduction and Related Work}
\label{sec:intro}

The complexity of modern financial reinforcement learning agents is such that individually innocuous actions may produce unexpectedly harmful results, and such failure modes can be difficult to detect or avoid.  One can postulate a practitioner who wishes to profit from an autonomous trading agent, but is aware that unconstrained objectives (e.g. ``construct as many paperclips as possible'') represent a risk to be thoughtfully managed, and wants to avoid some specific non-normative outcome.  \cite{bostrom2003ethical}  My specific aim for this article is to investigate the potential for \emph{unintentional} spoofing by an autonomous trading agent, and to introduce what I believe to be the first method to affirmatively inhibit that behavior during training.

\subsection{Reinforcement Learning}
\label{sec:intro_rl}

A Markov Decision Problem (MDP) can be thought of as a state machine in which the transitions are governed by a probability distribution conditioned on the action attempted by an agent.  \cite{bellman1957markov} After each action, the agent is assigned a numeric reward which it will attempt to maximize over the long term.

Reinforcement learning (RL) is the name given to a collection of approaches that solve MDPs when the problem definition must be discovered through experimentation with the environment. \cite{sutton2018reinforcement}  Model-free approaches to RL directly learn the overall utility of taking a certain action from a particular state.  Probably the most common is Q-learning, under which the agent requires minimal information: at each time step, it must be given a unique and consistent identifier for the current state of the environment and for each action that it might take.  \cite{watkins1992q}  The agent must use this information to learn the optimal action $a$ for each state $s$ by optimizing the Q-function:
\begin{equation}
\label{eq:q_function}
    Q^\pi(s,a) = R_s(a) + \gamma \sum_{s'} Pr_{ss'}[\pi(s)]V^\pi (s')
\end{equation}
which represents the sum of immediate and discounted expected future rewards that will result from doing $a$ while in $s$ and then continuing to follow policy $\pi$.  $R_s$ is the reward function for state $s$, $\gamma$ is the stepwise discount for future rewards, $Pr_{ss'}[\pi(s)]$ is the probability of a transition from state $s$ to new state $s'$ under policy $\pi$, and $V$ represents the utility or value of reaching a given state.

\subsection{Safe Reinforcement Learning}
\label{sec:intro_safe_rl}

Safe RL seeks to ensure that agents avoid catastrophic results by modifying their rewards or constraining their exploration of the state-action space.  For example, a trading agent's reward function could be penalized by the variance of returns over time to discourage excessive risk-taking, or a bipedal robot could be prohibited from exploring limb movements that would push its center of gravity past a tipping point.  Garcia collects many such approaches in an excellent survey.  \cite{garcia2015ACS}

Such techniques are effective for agents concerned with avoidance of immediately measurable consequences, but not well suited when failure modes:
\begin{itemize}
    \item are imprecisely defined, difficult to detect during training, or objectively similar to desired behavior;
    \item arise from sequences of individually acceptable actions;
    \item are extrinsic to the primary task at hand.
\end{itemize}
An example of a failure mode that fits all of these criteria would be the algorithmic violation of a financial regulation.  Many regulations reference intent, require interpretation, and result in penalties years later.  The current work considers exactly this challenging scenario.

\subsection {Normative Reinforcement Learning}
\label{sec:intro_normative_rl}

Normative RL targets activity patterns humans will find acceptable and appropriate to a situation, such as respecting personal space.  There is significant overlap with Safe RL, because normative behavior often implicates safety: an autonomous robot should not run over a human to more quickly reach its target location.  Soares et al. codified the \emph{value alignment problem}, which they define for an intelligent agent as learning and acting according to the preferences of its operators.  \cite{soares2014aligning}  Colloquially, we can imagine appending an implicit ``without doing anything bad'' to every objective.

A formal method for \emph{reward shaping} was introduced by Ng et al. in 1999 to guide a learning algorithm to discover the optimal policy more quickly by modifying the reward signal with additional feedback.  \cite{ng1999policy}  This was extended to \emph{policy shaping} by Griffith et al. in 2013, in which an agent attempts to directly learn the feedback as a separate policy, then combines the two policies to make decisions.  \cite{griffith2013policy}  These techniques are broadly applicable, but are considered foundations of normative RL, and I rely on both in my approach.

Normative RL has been studied by authors in robot navigation to follow the flow of pedestrian traffic, make long-term path predictions for cyclists, slow down at uncontrolled crossroads, or determine which member of a human group to approach with offers of assistance. \cite{okal2016learning}  Authors in narrative intelligence have explored learning human values by reading stories, then inducing desired mixtures of goal-oriented versus normative behaviors within interactive fiction game worlds.  \cite{riedl2016using,nahian2021training}  It is from this line of work that I draw the action reranking approach to policy shaping.

\subsection{Financial Market Spoofing}
\label{sec:intro_spoofing}

The Commodity Exchange Act (CEA), as modified by the Dodd-Frank Act, makes it unlawful to engage in order ``spoofing'', which it defines as ``bidding or offering with the intent to cancel the bid or offer before execution''.  In their guidance on the topic, the Commodity Futures Trading Commission (CFTC) lists some reasons a party might engage in spoofing, such as  \cite{cftc_spoofing_guidance}:
\begin{itemize}
    \item overloading a price quotation system;
    \item delaying another party's trade executions;
    \item creating the false appearance of supply or demand;
    \item creating artificial price movements upward or downward.
\end{itemize}
They also clarify that ``reckless trading, practices, or conduct will not constitute a `spoofing' violation''.  This may mean that inadvertent spoofing by an unconstrained intelligent trading agent would not currently be a violation, but as spoofing is identified as ``disruptive of fair and equitable trading'', I propose that we should codify best practices to avoid it.

Evidence of widespread spoofing with a price manipulation motive has been found in a custom data set provided by the Korea Exchange (KRX).  \cite{lee2013microstructure}  The CFA Institute's 2015 member survey reported that \emph{market fraud} and \emph{market trading practices} accounted for many members ``most serious ethical issue facing local market'': 49\% in China, 36\% in Japan, and 38\% each in the UK and US.  As summarized in a 2012 survey by Putni\c{n}\v{s}, theoretical and empirical economic literature has found spoofing to be profitable, implying harm to other traders from whom the excess profits must be extracted.  \cite{putnicnvs2012market}  In 2020 alone, the CFTC filed and resolved 16 spoofing cases, including a record \$920 million settlement. \cite{cftc2020spoofing}

\subsection{Spoofing Detection}
\label{sec:intro_detection}

Detection of spoofing is a topic of common interest.  Cao et al. described two forms of deceptive electronic price manipulation.  \cite{cao2014detecting}  In \emph{spoofing trading}, an illusion of demand is generated over a relatively long time horizon by maintaining a large order on the opposite side of the order book.  In \emph{quote stuffing}, the illusion of a bidding war is generated by placing a sequence of orders inside the bid-ask spread.  The authors transform historical data for four stocks to relative price offsets, then inject synthetic spoofing sequences.  Using K-Nearest Neighbors and Support Vector Machines, they detect the spoofing sequences with positive results.  My detection approach is similar, but I use an agent-based simulation for the entire market to ensure that other traders can react to any spoofing behavior.

Leangarun et al. found that a feedforward neural network with access to only Level 1 data (i.e. successful transactions) could detect a classic ``pump and dump'', but could not detect spoofing.  \cite{leangarun2016stock}  Mendon{\c{c}}a et al. obtained private data from a Brazilian brokerage firm with a consistent (but anonymous) identifier for each trader, and found that an expert-designed decision tree could identify likely spoofing behavior at the brokerage level.  \cite{mendoncca2020detection}  Li et al. were able to detect days on which spoofing had occcured using a variety of classification methods on data obtained from Chinese regulators after publicized enforcement actions. \cite{li2017market}  Wang et al. used Generative Adversarial Networks (GAN) to train both a spoofing agent which learns to avoid detection by disguising its activity as market making, and a detector that attempts to defeat this evasion. \cite{wang2020market}

\subsection{Inadvertent Spoofing}
\label{sec:intro_inadvertent}

Suppose that a responsible practitioner of financial ML undertakes to create an RL-based stock trading agent with an input space consisting of technical market features and sufficient internal state to render the problem Markovian.  The output space will be designed to permit placing or cancelling orders to buy or sell stock at various offsets to the best available pricing at the time.  The practitioner hopes that the agent will discover a profitable trading strategy.

The deployment of ML and RL-based trading agents presents a potential new universe of regulatory difficulties.  Under the guidance discussed in \Cref{sec:intro_spoofing}, it seems impossible for a human trader to \emph{inadvertently} spoof a market, but it is entirely plausible that our practitioner's intelligent agent will discover that spoofing is the optimally-profitable strategy.  I aim to help the practitioner affirmatively avoid that outcome with my primary contributions: demonstrating that such an unconstrained agent \emph{can} inadvertently learn to spoof and introducing a potential solution that should generalize to other unwanted behaviors as well.

\section{Approach}
\label{sec:approach}

My approach to avoid learning behaviors for which there is no programmable definition (e.g. spoofing) requires a behavior recognizer.  To learn such a recognizer requires action sequences labeled as containing or not containing the behavior.  Therefore, although spoofing detection is not a primary focus of this work, it is a necessary starting point.  The trained detector will form the basis of my method to deter inadvertent spoofing by the RL agent.

\subsection{Spoofing Data Synthesis}
\label{sec:approach_simulation}

Spoofing is a behavior that relies on sequential, not individual, actions.  Due to the aggregated and anonymous nature of public market data, no two orders can be traced to the same actor, meaning sequences of actions cannot be identified.  To ensure relevant action sequences can be identified, all experiments are performed within a common open source financial market simulation.  \cite{byrd2020abides}

I first instantiate a population of stylized trading agents that pursue various greedy but legal strategies.  I then introduce a deliberately designed spoofing agent.  It observes the simulated market to understand current prices, then purchases some stock at a ``fair'' price:
\begin{equation}
    \mathbb{E}[p_e] \le \min \big[ \frac{a_w^L + a_w^H}{2},  p'_e \big]
\end{equation}
where $\mathbb{E}[p_e]$ is the expected entry price for its desired position, $p'_e$ is the prior entry price, and $a_w^L$ and $a_w^H$ are respectively the lowest and highest observed \emph{best ask} prices during the warmup period.  The agent has configurable spoofing order quantity and depth, and follows this general strategy:
\begin{enumerate}
    \item Buy (or already own) shares of ABC stock.
    \item Place ABC limit bids slightly under the current best bid.
    \item Wait for other traders to bid up the price after observing the apparent demand.
    \item Adjust spoofing bids to remain near the best bid, but do not allow execution.
    \item Sell the ABC position and cancel the spoofing orders.
\end{enumerate}

Within each simulated market day, every order-related action is recorded: agent id, agent type, timestamp, action type (order/cancel), direction (buy/sell), limit price relative to best bid/ask, quantity, spoofing label.  After simulation, an action sequence is reconstructed for each agent for each market day, then divided into non-overlapping subsequences of twenty actions each.  Four features are retained in the training examples: action type, order direction, relative limit price, and order quantity.  Examples are labeled True when they arise from a spoofing agent not configured in its ``honest'' mode, and False otherwise.  Each training example has dimensionality $(20,4)$.

\subsection{Spoofing Detection}
\label{sec:approach_detection}

The temporal element of spoofing means that rearranging the same action primitives can render a spoofing explanation likely or impossible, as in the example below, where $M^+$ indicates a market buy, $L^+$ indicates a limit buy under the best bid, $C$ indicates to cancel one $L^+$, and $M^-$ indicates a market sell.

\begin{center}
\begin{tabular}{rcccccccc}
\midrule
Spoofing & $M^+$ & $L^+$ & $L^+$ & $L^+$ & $M^-$ & $C$ & $C$ & $C$ \\
Not spoofing & $L^+$ & $C$ & $M^-$ & $L^+$ & $L^+$ & $M^+$ & $C$ & $C$ \\
\bottomrule\vspace{-3mm}
\end{tabular}
\end{center}

This characteristic suggests an \emph{activity recognition} problem and the need for a learning method that assigns value to the temporal ordering of the data.  Lara et al. present a survey of approaches using decision trees, instance-based learners, and artificial neural networks \cite{lara2012survey}; and Wang et al. survey approaches using convolutional neural networks (CNN), autoencoders, and recurrent neural networks (RNN).  \cite{wang2019deep}  I evaluate detection using a variety of neural network architectures and refer to the learned detector as a real-valued function $\Theta(a_0:a_{19})\rightarrow [0,1]$.


\begin{figure*}[t]

\includegraphics[width=\textwidth]{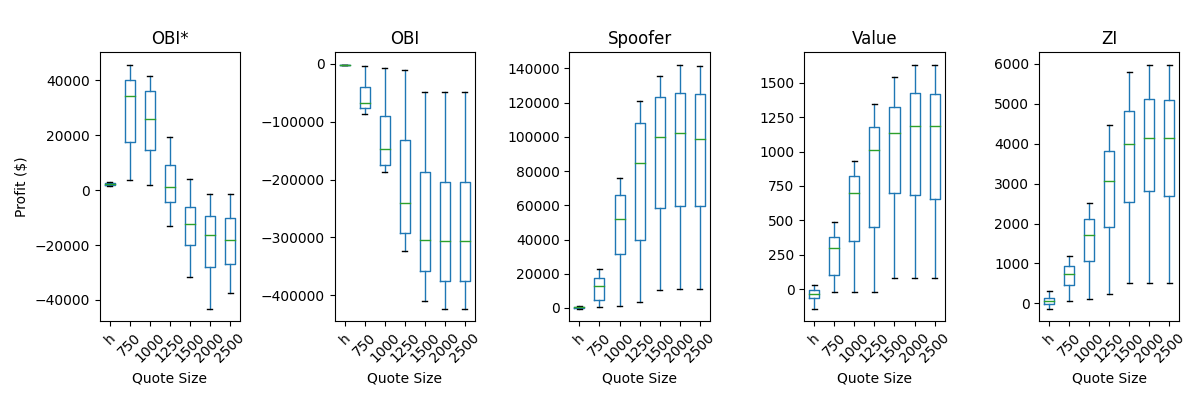}
\caption{Effect of initial spoofing agent quote size on profitability of each agent class.  Includes all quote depths.}
\label{fig:detect_spoof_quote_size}
\end{figure*}

\subsection{Trading with Fixed Policies}
\label{sec:approach_fixed_policies}

The agent and environment are validating by ensuring that profit is possible under the state-action space described in \Cref{sec:approach_normative}.  For this I employ a pair of fixed policies: $\pi^{s}$, which will attempt to profit by spoofing the simulated market, and $\pi^{h}$, which will attempt to profit without spoofing the simulated market.  If both policies are profitable, this will demonstrate that an RL agent using this state-action space \emph{could} learn to profit either through spoofing or honest participation.

Honest policy $\pi^h(s) \rightarrow a$ is constructed as:
\begin{equation}
    \pi^h(s) =
    \begin{cases*}
        \text{AG} & if $\lnot$ H $\land$ EA $\land$ $\lnot$ OP \\
        \text{EX} & if H $\land$ XA $\land$ $\lnot$ OP \\
        \text{CN} & if OP \\
        \text{DN} & otherwise \\
    \end{cases*}
\end{equation}

Spoofing policy $\pi^s(s) \rightarrow a$ is constructed as:
\begin{equation}
    \pi^s(s) =
    \begin{cases*}
        \text{AG} & if $\lnot$ H $\land$ EA $\land$ $\lnot$ OP \\
        \text{PS} & if H $\land$ $\lnot$ (XA $\lor$ SL) $\land$ $\lnot$ OP \\
        \text{EX} & if H $\land$ (XA $\lor$ SL) \\
        \text{CN} & if $\lnot$ H $\land$ OP \\
        \text{UP} & if H $\land$ $\lnot$ (XA $\lor$ SL) $\land$ DP \\
        \text{DN} & otherwise \\
    \end{cases*}
\end{equation}

\subsection{Trading with Normative Alignment}
\label{sec:approach_normative}

I construct a long-only trading agent that interacts with the simulated market to optimize \Cref{eq:q_function}.  The agent is constrained to hold a set quantity of stock to avoid conflation of intelligent behavior with mere leverage.  Its base reward function includes the realized gain/loss in dollars when a stock position is closed and a small transaction cost for every request sent to the stock exchange.  I investigate two exploration methods: $\epsilon$-greedy, in which $\epsilon \in [0,1]$ controls the proportion of random vs expected-optimal actions selected; and Boltzmann (softmax), in which actions are stochastically selected from a Boltzmann distribution \cite{watkins1989learning}:
\begin{equation}
\label{eq:boltzmann}
P(a|s)=\frac{e^{Q_{sa}}}{\sum_j e^{Q_{sj}}}
\end{equation}

The current state of the environment (agent included) is represented as a set of boolean conditions drawn from the following:
\begin{itemize}
    \item \textbf{Holdings} (H): is the agent in a long position?
    \item \textbf{Holdings Advantage} (HA): did the agent have entry advantage when the current position was acquired?
    \item \textbf{Entry Advantage} (EA): is the estimated cost to enter a position less than the observed reasonable price?
    \item \textbf{Exit Advantage} (XA): does the estimated return from exiting a position exceed the agent's profit target?
    \item \textbf{Stop Loss} (SL): does the estimated loss from exiting a position exceed the agent's maximum permitted loss?
    \item \textbf{Open Orders} (OP): does the agent have open, unfilled orders?
    \item \textbf{Depth Orders} (DP): has the market moved since the agent placed an unfilled order?
\end{itemize}

Each experimental agent is given an action space drawn from the following:
\begin{itemize}
    \item \textbf{Aggressive Entry} (AG): Aggressively enter a long position by purchasing a set quantity of stock at whatever price is immediately available.
    \item \textbf{Passive Entry} (PS): Passively enter a long position by offering to buy a set quantity of stock at slightly below the current price.
    \item \textbf{Exit} (EX): Exit a long position by selling whatever quantity of stock is owned.
    \item \textbf{Cancel} (CN): Cancel all unfilled open orders.
    \item \textbf{Update Passive Entry} (UP): Refresh a passive entry attempt by cancelling open orders and reissuing a passive entry order.  This effectively combines action CN with PS.
    \item \textbf{Do Nothing} (DN): Do nothing.
\end{itemize}

For experiments with normative guidance using detector $\Theta(a_0:a_{19})$, all price and quantity inputs are scaled according to the detector's training data.  Action sequences are continued across market days for the same agent, and the first 20 total actions of each agent are assumed to be normative.

For experiments with reward shaping, the reward for closing a profitable position is transformed by the activation of spoofing detector $\Theta(a_0:a_{19})$:
\begin{equation}
\label{eq:shaped_rewards}
    r' = r \times [1 - \Theta(a_0:a_{19})]
\end{equation}

For experiments with action reranking, the Boltzmann-derived action probability vector from \Cref{eq:boltzmann} is transformed prior to action selection.  I tentatively add each candidate action to recent history and use the $\Theta(a_0:a_{19})$ activation of the ``proposed'' history to estimate the contextual normativity of each action, then update the probability to select each action:
\begin{equation}
\label{eq:reranked_actions}
    P(a|s)=\frac{e^{Q_{sa} \cdot [1 - \Theta(a,a_0:a_{18})]}}{\sum_j e^{Q_{sj} \cdot [1 - \Theta(j,a_0:a_{18})]}}
\end{equation}

\section{Experiments and Results}
\label{sec:experiments}

A sequence of experiments and their results are presented together for logical consistency.  The results of earlier experiments influence the design of later experiments.

\subsection{Initial Spoofing Strategy}
\label{sec:exp_adhoc}

To achieve the ultimate aim of learning not to spoof, I must first ensure that spoofing our simulated market is both possible and profitable.  I instantiate a population of: 500 Zero Intelligence (ZI) agents, which submit bids and offers drawn from a stochastic distribution around a private extrinsic valuation; 500 Value agents which arbitrage the current market price against a private extrinsic valuation; and 10 Order Book Imbalance (OBI) agents, which represent high-frequency liquidity traders predicting short term price moves by studying the order book.  Value and ZI agents are randomly geolocated, OBI agents have communication latency $[21,399] {\mu}s$, and the experimental agent is always exchange co-located with latency $33 ns$.  In figures, the label \textbf{Exper.} indicates the experimental agent, and \textbf{OBI*} indicates the high frequency agent nearest the exchange.

The efficacy of the spoofer and its effect on other agents are tested by simulating a total of 1,160 full market days with varied spoofing hyperparameter configurations.  One configuration, labeled $h$ in plots, includes the spoofing agent configured in an ``honest'' mode that omits the manipulative portion of its strategy.  Each spoofing configuration is simulated for 20 market days; the non-spoofing configuration is simulated for 200 days.  The effect of spoofing quote size on the profit distribution of each agent is summarized in \Cref{fig:detect_spoof_quote_size}.  From the results, I conclude:
\begin{itemize}
    \item The order book aware agents (OBI) are harmed by spoofing.  OBI* is harmed least.  Harms increase with spoofing order quantity up to a limit.  Quote depth is irrelevant within the tested range  of 3 to 8 (plot omitted due to page limit).
    \item Value and ZI agents are indirectly helped by spoofing, presumably because the OBI agents had been exploiting \emph{them} and now cannot.
    \item The spoofing strategy is clearly effective regardless of precise parameter selection, but lower quote sizes engender less response from the OBI agents, and hence produce lower spoofing profits.
\end{itemize}
As the strategy is profitable and shows expected results, it should be suitable to synthesize data for the spoofing detector.


\begin{figure*}[t]

\includegraphics[width=\textwidth]{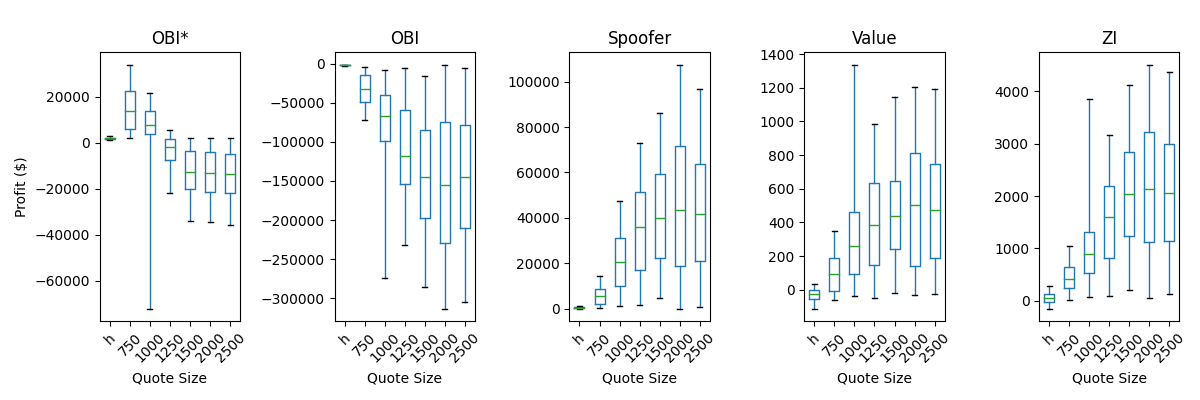}
\caption{Effect of fixed policy agent quote size on profitability of each agent class.  Includes all quote depths.}
\label{fig:no_spoof_fixed_quote_size}
\end{figure*}


\begin{table}[b]
\centering
\captionsetup{justification=centering,width=\linewidth}
\caption {Spoofing detector feature ablation.}
\begin{tabular}{c|rrrrr}
Features & FP & FN & Precision & Recall & MCC \\
\midrule
D & 3,411 & 18 & 0.832 & 0.999 & 0.909 \\
P & 294,816 & 7,055 & 0.019 & 0.583 & 0.000 \\
Q & 0 & 15,537 & 1.000 & 0.082 & 0.283 \\
T & 556 & 18 & 0.968 & 0.999 & 0.983 \\
\midrule
DP & 3,247 & 20 & 0.839 & 0.999 & 0.912 \\
DQ & 3,699 & 20 & 0.824 & 0.999 & 0.903 \\
DT & 88 & 4 & 0.995 & 1.000 & 0.997 \\
PQ & 0 & 15,511 & 1.000 & 0.084 & 0.285 \\
PT & 567 & 18 & 0.968 & 0.999 & 0.983 \\
QT & 554 & 25 & 0.969 & 0.999 & 0.983 \\
\bottomrule
\end{tabular}
\label{tab:detect_spoof_detection_ablation}
\end{table}

\subsection{Spoofing Detector}
\label{sec:exp_detection}

The data synthesis described in \Cref{sec:approach_simulation} results in an imbalanced labeled data set of size $(2611707, 20, 4)$ with 84,666 (approx. 3.2\%) examples of spoofing action sequences.  The examples are shuffled and divided into training (64\%), validation (16\%), and holdout test (20\%) sets.  Each set is required to maintain the overall class balance of labels.  The action type and order direction features are converted from boolean to 0-1 categorical.  The price and quantity features are normalized to $N(0,1)$.  Class imbalance is corrected by oversampling the positive class.

The task is activity recognition (see \Cref{sec:approach_detection}) and I compare six network architectures which may capture the required temporal patterns: bidirectional and standard Long Short-Term Memory (Bi-LSTM, LSTM), bidirectional and standard Gated Recurrent Units (Bi-GRU, GRU), a CNN convolving over time, and a simple feed-forward neural network (FFNN).  Each network is similarly configured with the experimental layer accepting input shape $(20,4)$ with 64 hidden units and ReLu activation, connected to a dense layer with sigmoid activation and a single output neuron, and trained with batch size 64 using the Adam optimizer and binary crossentropy loss function.


\begin{table}[b]
\centering
\captionsetup{justification=centering,width=\linewidth}
\caption {Spoofing detector architecture comparison.}
\begin{tabular}{cl|rrrrrrr}
Feat. & Arch. & FP & FN & Precision & Recall & MCC \\
\midrule
DT & FFFC & 125 & 6 & 0.993 & 1.000 & 0.996 \\
 & CNN & 36 & 1 & 0.998 & 1.000 & 0.999 \\
 & GRU & 69 & 0 & 0.996 & 1.000 & 0.998 \\
 & LSTM & 97 & 4 & 0.994 & 1.000 & 0.997 \\
 & Bi-GRU & 32 & 0 & 0.998 & 1.000 & 0.999 \\
 & Bi-LSTM & 88 & 4 & 0.995 & 1.000 & 0.997 \\
\bottomrule\vspace{-3mm}
\end{tabular}
\label{tab:detect_spoof_detection_model}
\end{table}

Every combination of the available features (order \textbf{D}irection, relative \textbf{P}rice, order \textbf{Q}uantity, action \textbf{T}ype) is empirically tested against each of the proposed network architectures with five-fold cross validation.  In \Cref{tab:detect_spoof_detection_ablation}, I present the results of a feature ablation study using out-of-sample results for the Bi-LSTM architecture as a representative model.  Combinations of three or four features do not improve accuracy and are omitted for brevity.  The table reports false positives and negatives, precision, recall, and the Matthews Correlation Coefficient (MCC) for imbalanced binary classification.  Action type and order direction are individually very strongly correlated with ground truth.  I use their combination with $MCC=0.997$ as the final features for spoofing detection.

The six different model architectures presented in \Cref{sec:approach_detection} are evaluated on the combination of features selected by the study: order direction and action type.  The results are presented in \Cref{tab:detect_spoof_detection_model}.  Using the selected features in the simulated market, even the simpler architectures succeed at the detection task.  I posit this is possible because the length 20 input data windowing incorporates time within each example behavior.  The final detector will be the temporally-convolved CNN with features action type and order direction.


\begin{figure*}[t]
\includegraphics[width=\textwidth]{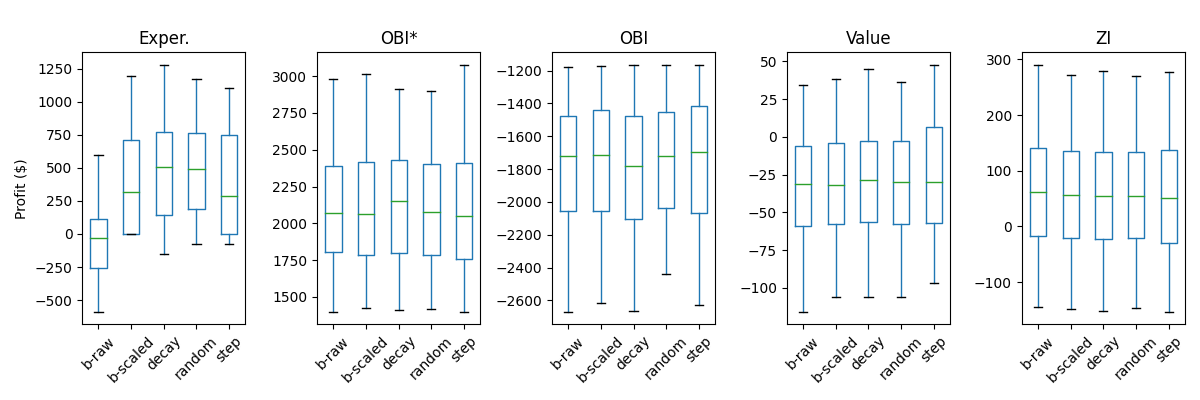}
\caption{Effect of restricted Q-trader exploration method on profitability of each agent class.}
\label{fig:no_spoof_learn_honest_profit}
\end{figure*}

\subsection{Fixed Policy Agents}
\label{sec:exp_fixed_policies}

To ensure meaningful results are possible in the RL experiments, I first evaluate the fixed policy agents $\pi^s(s)$ and $\pi^h(s)$ defined in \Cref{sec:approach_fixed_policies}.  If both policy agents are profitable, this will demonstrate that a similar RL agent could learn to profit either with or without spoofing the market.  Agent $\pi^s(s)$ or $\pi^h(s)$ replaces the initial spoofing agent; the environment is otherwise identical to \Cref{sec:exp_adhoc}.  \Cref{fig:no_spoof_fixed_quote_size} summarizes the daily profit of each strategy across 100 simulated full market days.  Quote size $h$ represents experiments with $\pi^h(s)$; all other quote sizes use $\pi^s(s)$.  Policy $\pi^h(s)$ outperforms the original honest strategy, while $\pi^s(s)$ underperforms the original spoofing strategy, but returns for both policies are positive and of the same order of magnitude as the original strategies.

I record the average activation of the spoofing detector for both policies and find that $\Theta(\pi^h(s),a_0:a_{19})=0.000$ and $\Theta(\pi^s(s),a_0:a_{19})=0.842$.  The detector ``flags'' the spoofing policy and not the honest policy, which is expected.  Since both policies demonstrate consistent profit, I conclude that an RL agent with similar state-action space could learn to profit in the market without spoofing, but that adopting spoofing would enhance profitability.


\begin{table}[b]
\centering
\captionsetup{justification=centering,width=\linewidth}
\caption {Profit by agent class by $Q^R$ exploration approach.}
\begin{tabular}{l|rrrrr}
Exploration & $Q^R$ & OBI & OBI* & Value & ZI \\
\midrule
$\epsilon$-greedy random & 467 & -1,761 & 2,112 & -30 & 57 \\
$\epsilon$-greedy step & 382 & -1,787 & 2,118 & -27 & 54 \\
$\epsilon$-greedy decay & 482 & -1,825 & 2,138 & -30 & 58 \\
Boltzmann $r$-raw & -58 & -1,791 & 2,117 & -34 & 62 \\
Boltzmann $r$-scaled & 405 & -1,776 & 2,123 & -32 & 59 \\
\bottomrule
\end{tabular}
\label{tab:no_spoof_learn_honest_profit_test}
\end{table}

\subsection{Restricted RL: Unable to Spoof}
\label{sec:exp_restricted}

I now replace the fixed policy agent with a restricted agent $Q^R$ which is given the complete state space described in \Cref{sec:approach_normative} but limited to action space \{AG, EX, DN\}.  It adopts online Q-Learning (see \Cref{sec:intro_rl}) to discover a trading policy that maximizes profit.  The agent can transact in the market but has no ability to cancel orders.  I hypothesize that $Q^R$ should be unable to learn the non-normative spoofing behavior.

Multiple variations of $Q^R$ are tested: three using $\epsilon$-greedy exploration (see \Cref{sec:approach_normative}) with all random actions, a scheduled step decline per market day, or a more typical geometric decay; two with Boltzmann exploration (see \Cref{sec:approach_normative}) receiving unscaled or linearly scaled environmental rewards.  Ten separate random experiments are run for each variation.  Every agent is permitted to train for 10 full market days, then evaluated for an additional 10 full market days.  A slight initial bias is given to action DN in all cases. Learning rate decay is applied per state-action pair: $\alpha=\mathrm{max}(0.1,\frac{1}{c(s,a)})$, where $c$ is a counter function.

\Cref{tab:no_spoof_learn_honest_profit_test} and \Cref{fig:no_spoof_learn_honest_profit} present the mean and distribution of profits across 100 market days for each variation of $Q^R$.  All variations are profitable in the mean except unscaled-$r$ Boltzmann; I postulate that receiving early large rewards may skew its action selection.  The impact on other trading agents is similar across $Q^R$ variations.  The Boltzmann agent converges in fewer days of training; I postulate that it more quickly abandons unrewarding actions.  Subsequent experimental agents are based on the scaled-$r$ Boltzmann variation of $Q^R$, due to its fast convergence, near-optimal profit, and lack of significant downward profit deviations.

The average activation of spoofing detector $\Theta(a_0:a_{19}) < 1e-17$ for all variations of $Q^R$, supporting the hypothesis that the restricted agent lacks the capacity to spoof the simulated market.  Without spoofing, $Q^R$ consistently learns to profit: in the scaled-$r$ Boltzmann variation, only 5\% of market days end in a loss, and the average loss on those days is only \$195.  I conclude that a properly configured RL trader can learn to profit in the market without spoofing.

\begin{figure*}[t]
\includegraphics[width=\textwidth]{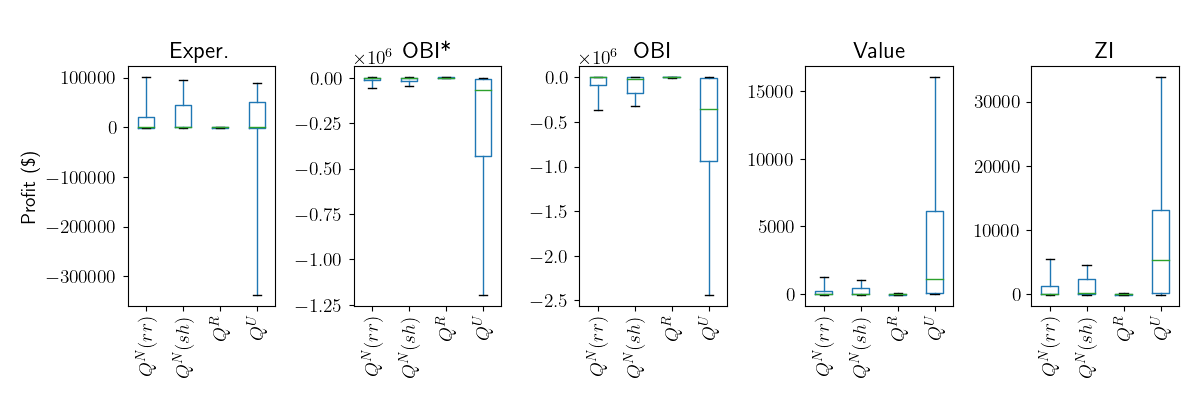}
\caption{Effect of normative guidance method on profitability of each agent class.  Previous Q-traders included for reference.}
\label{fig:no_spoof_no_spoof_profit}
\end{figure*}

\subsection{Unconstrained RL: Learning to Spoof}
\label{sec:exp_unconstrained}

Through prior experiments, we have seen that a typical spoofing strategy and a spoofing policy agent can successfully spoof the simulated market.  We have also seen that an honest policy agent and an action-restricted RL trader can profit in the same market without spoofing.  Before introducing normative guidance, I investigate what happens without it.

Restricted agent $Q^R$ is replaced with an unconstrained Q-Learning agent $Q^U$ which has access to the complete state-action space described in \Cref{sec:approach_normative} and attempts to maximize profit.  No effort is made to guide or shape the learning agent towards spoofing; its behavior is merely observed.  I hypothesize that $Q^U$ will learn to maximize profit by spoofing the market.  All experiments are conducted with the same market configuration, training process, and scaled-$r$ Boltzmann variant from \Cref{sec:exp_restricted}.  Order depth was previously found to have no effect on spoofing outcomes, so all experiments are conducted with fixed depth 5 for action PS.

\Cref{tab:exp_unconstrained} presents the mean profit for each agent class across all out of sample simulations.  Rows labeled \emph{init} and $\pi^s(s)$ respectively present the initial and fixed policy spoofing strategies for comparison.  In the mean, unconstrained agent $Q^U$ obtains greater profit than the reference spoofing strategies for most tested quantities.  The distribution of returns for $Q^U$ is presented in \Cref{sec:exp_normative} alongside the normative agents for comparison.

Mean spoofing detector activation $\Theta(a_0:a_{19}) = 0.755$ across all quote sizes for agent $Q^U$.  Agent $Q^U$ profits at levels similar to the known spoofing strategies, and well in excess of the known honest strategies.  As with the known spoofing strategies, $Q^U$ primarily profits from the high-frequency OBI traders.  I conclude that agent $Q^U$ has inadvertently learned to maximize profit by spoofing the simulated market, and that normative guidance may indeed be necessary to avoid such behavior.


\begin{table}[b]
\centering
\captionsetup{justification=centering,width=\linewidth}
\caption {Profit by agent class by $Q^U$ passive order quantity.}
\begin{tabular}{c|rrrrr}
Quantity & $Q^U$ & OBI & OBI* & Value & ZI \\
\midrule
\emph{init} & 88,985 & -267,769 & -15,494  & 1,011 & 3,661 \\
$\pi^s(s)$ & 48,627 & -158,853 & -14,827 & 531 & 2,261 \\
\midrule
750 & 23,644 & -118,565 & 63,851 & 466 & 1,493 \\
1,000 & 100,450 & -293,520 & 20,455 & 1,223 & 3,818 \\
1,250 & 119,480 & -499,619 & -73,265 & 1,875 & 7,025 \\
1,500 & 120,997 & -546,697 & -101,012 & 2,068 & 7,732 \\
2,000 & 96,195 & -614,292 & -143,260 & 2,523 & 8,627 \\
2,500 & 61,764 & -756,587 & -214,718 & 3,413 & 10,510 \\
\bottomrule
\end{tabular}
\label{tab:exp_unconstrained}
\end{table}

\subsection{Normative RL: Learning Not to Spoof}
\label{sec:exp_normative}

In \Cref{sec:exp_unconstrained} we saw that an unconstrained Q-Learning agent tasked with simple profit maximization may learn to spoof a financial market in which it participates.  In the final set of experiments, I employ spoofing detector $\Theta(a_0:a_{19})$ as a source of normative guidance, aiming to prevent the discovery of spoofing as an ``optimal'' strategy.


\begin{table}[b]
\centering
\captionsetup{justification=centering,width=\linewidth}
\caption {Profit by agent class by experimental configuration.}
\begin{tabular}{c|r|rrrrr}
Config. & $\Theta(a)$ & Exper. & OBI & OBI* & Value & ZI \\
\midrule
$\pi^h(s)$ & 0.000 & 517 & -1,654 & 1,985 & -30 & 55 \\
$\pi^s(s)$ & 0.842 & 48,627 & -158,853 & -14,827 & 531 & 2,261 \\
$Q^R$ & 0.000 & 405 & -1,776 & 2,123 & -32 & 59 \\
$Q^U$ & 0.755 & 61,764 & -756,587 & -214,718 & 3,413 & 10,510 \\
\midrule
$Q^N(rr)$ & 0.043 & 21,871 & -83,964 & -12,070 & 227 & 1,263 \\
$Q^N(sh)$ & 0.072 & 19,987 & -86,130 & -10,882 & 263 & 1,268 \\
\bottomrule
\end{tabular}
\label{tab:normative_profit}
\end{table}

I replace unconstrained agent $Q^U$ with $Q^N$, a Q-Learning agent which maintains access to the full state-action space from \Cref{sec:approach_normative}.  Agent $Q^N(sh)$ receives normative guidance through reward shaping; agent $Q^N(rr)$ receives normative guidance through action reranking.  Both methods are shown in \Cref{sec:approach_normative}. Experimental parameters are identical to \Cref{sec:exp_unconstrained} except as noted.

All available actions are individually normative, and detector $\Theta(a_0:a_{19})$ was trained on action sequences of length 20 as visible from the exchange.  I therefore maintain a length 20 history per agent of exchange-visible order activity.  This is distinct from the agent's internal selections from the action space.  The use of this history is explained in \Cref{sec:approach_normative}.  I hypothesize that normative trader $Q^N$ will use the additional flexibility of its action space to outearn restricted agent $Q^R$, but with lower profit and spoofing detector $\Theta(a_0:a_{19})$ activation than unconstrained agent $Q^U$.  

\Cref{tab:normative_profit} presents the mean profit for each agent class.  Policy and Q-Learning agents from prior experiments are presented for comparison.  Column $\Theta(a)$ contains the mean activation of the spoofing detector for the experimental agent's actions.  Normative agent $Q^N$ achieves approximately forty times the profit of prior experimental non-spoofing agents.  It achieves 30-40\% of the profit of prior experimental spoofing agents with an order of magnitude lower activation of spoofing detector $\Theta(a_0:a_{19})$.

\Cref{fig:no_spoof_no_spoof_profit} presents the distribution of profits for each agent class.  Unconstrained agent $Q^U$ experiences infrequent but catastrophic losses; I posit these occur when spoofing limit orders are accidentally transacted.  Both variations of $Q^N$ eliminate those losses. Compared to normative agent $Q^N$, unconstrained agent $Q^U$ also inflicts far greater volatility on the returns of other traders.

Stipulating that in the present work, I equate spoofing with having a high activation of spoofing detector $\Theta(a_0:a_{19})$, and the efficacy of the detector is therefore critical, I observe of the Q-Learning trading agents that:
\begin{itemize}
    \item Restricted agent $Q^R$ is unable to spoof and attains limited profit.
    \item Unconstrained agent $Q^U$ learns to spoof and attains very high mean profit, but causes high variance in the profitability of all agents including itself.
    \item Normative agent $Q^N$ attains high mean profit without learning to spoof, and is less disruptive to the overall market.
    \item Action reranking produces lower $\Theta(a_0:a_{19})$ activation and slightly higher mean profit than reward scaling, but with fewer upward profit deviations.
\end{itemize}
I conclude that both studied forms of normative guidance capture the increased profit potential of an RL-based trading mechanism, but with much lower activation of spoofing detector $\Theta(a_0:a_{19})$ and much less disruption to other market participants.  This could offer a new direction for regulators to explore: encouraging or requiring best practices that include such normative guidance to curtail the inadvertent adoption of non-normative or disruptive practices by autonomous RL-based trading agents.

\section{Conclusion}

I investigated the risk of an honest practitioner of financial machine learning inadvertently producing a Q-Learning trading agent that spoofs the market in which it participates.  First a spoofing recognizer was learned using variations of a hand-designed spoofing agent in a simulated market, showing that simulation is a safe and effective approach to synthesize examples of non-normative behavior.  I found that spoofing was detectable in the market and that order-book aware agents are harmed by spoofing as expected.  I introduced and tested honest and spoofing fixed policy trading agents, action-restricted and unconstrained Q-trading  agents, and flexible Q-trading agents with two forms of normative guidance.  The relative profits and spoofing detector activation levels suggest that the restricted Q-trader could not spoof, the unconstrained Q-trader inadvertently learned to spoof, and the Q-traders with normative guidance achieved respectable profits without learning to spoof.  Both approaches to normative guidance allow profitability with less market disruption and the avoidance of apparent spoofing behavior.  I acknowledge that the normative guidance can only inhibit behavior that it recognizes; detector robustness should be a high priority.  

More broadly, I believe that unintentional legal or regulatory violations by intelligent trading algorithms are a serious concern requiring immediate attention, and conclude with a few recommendations: Practitioners could employ an ensemble of non-normative behavior detectors to better guide autonomous agents.  Regulatory agencies could recommend or mandate that firms deploy such an ensemble as a best faith effort to detect and diminish accidentally-disruptive behavior by their autonomous intelligent trading agents.  The use of market simulation to understand how other traders may react to an agent's behavior could inform the development of component models for such an ensemble of normative guides.

\begin{acks}
This material is based upon research supported by the National Science Foundation under Grant No. 1741026 and by a JP Morgan Fellowship.
\end{acks}

\bibliographystyle{ACM-Reference-Format}
\bibliography{spoofing}

\end{document}